\documentclass[letterpaper, 10 pt, conference]{ieeeconf}  
\IEEEoverridecommandlockouts                              

\usepackage[utf8]{inputenc}
\usepackage{wasysym}
\usepackage{amssymb} 
\usepackage{comment}
\usepackage{graphics} 
\usepackage{epsfig} 
\usepackage{siunitx}
\usepackage[official]{eurosym}
\usepackage{csquotes}
\usepackage{hyperref}
\usepackage{url} 
\usepackage[capitalise]{cleveref}
\usepackage{fontawesome}
\let\labelindent\relax
\usepackage{enumitem}
\usepackage{booktabs}
\usepackage{multirow}
\usepackage[table,dvipsnames]{xcolor}

\definecolor{mygray}{gray}{.95}

\usepackage{placeins}

\usepackage{subcaption}
\DeclareCaptionFormat{twolinetab}{#1#2#3}
\usepackage{tabularx}
\usepackage{graphicx}
\usepackage{adjustbox}
\usepackage{array}

\usepackage{makecell}

\newcommand{\metrabs}[0]{MeTRAbs}
\newcommand{\crazypose}[0]{\emph{FISHnCHIPS}}

\title{\LARGE \bf
Systematic Comparison of Projection Methods for \\
Monocular 3D Human Pose Estimation on Fisheye Images}

\author{Stephanie Käs$^{1}$, Sven Peter$^{1}$, Henrik Thillmann$^{1}$,
        \\
        Anton Burenko$^{1}$,
        David Benjamin Adrian$^{2}$, Dennis Mack$^{2}$, Timm Linder$^{2}$,
        Bastian Leibe$^{1}$%
        \thanks{$^{1}$ Chair for Computer Vision, RWTH Aachen University, Germany. Mail: 
                \textrm{\{lastname\}@vision.rwth-aachen.de}              
                \newline
                $^{2}$ Robert Bosch GmbH, Corporate Research \& Bosch Center for AI, Renningen and Hildesheim, Germany. Mail: 
                \textrm{\{firstname.lastname\}@de.bosch.com}
        }
}
        
\begin{document}
\maketitle
\thispagestyle{empty}
\pagestyle{empty}

\begin{abstract}
Fisheye cameras offer robots the ability to capture human movements across a wider field of view (FOV) than standard pinhole cameras, making them particularly useful for applications in human-robot interaction and automotive contexts. However, accurately detecting human poses in fisheye images is challenging due to the curved distortions inherent to fisheye optics. While various methods for undistorting fisheye images have been proposed, their effectiveness and limitations for poses that cover a wide FOV has not been systematically evaluated in the context  of absolute human pose estimation from monocular fisheye images. To address this gap, we evaluate the impact of pinhole, equidistant and double sphere camera models, as well as cylindrical projection methods, on 3D human pose estimation accuracy. We find that in close-up scenarios, pinhole projection is inadequate, and the optimal projection method varies with the FOV covered by the human pose. The usage of advanced fisheye models like the double sphere model significantly enhances 3D human pose estimation accuracy. 

We propose a heuristic for selecting the appropriate projection model based on the detection bounding box to enhance prediction quality.

Additionally, we introduce and evaluate on our novel dataset \crazypose{}, which features 3D human skeleton annotations in fisheye images, including images from unconventional angles, such as extreme close-ups, ground-mounted cameras, and wide-FOV poses, available at: \\
{\color{Blue}\href{https://www.vision.rwth-aachen.de/fishnchips}{https://www.vision.rwth-aachen.de/fishnchips}}
\end{abstract}


\section{Introduction}
\begin{figure}[t!]  
    \centering
    \includegraphics[width=0.9\linewidth]{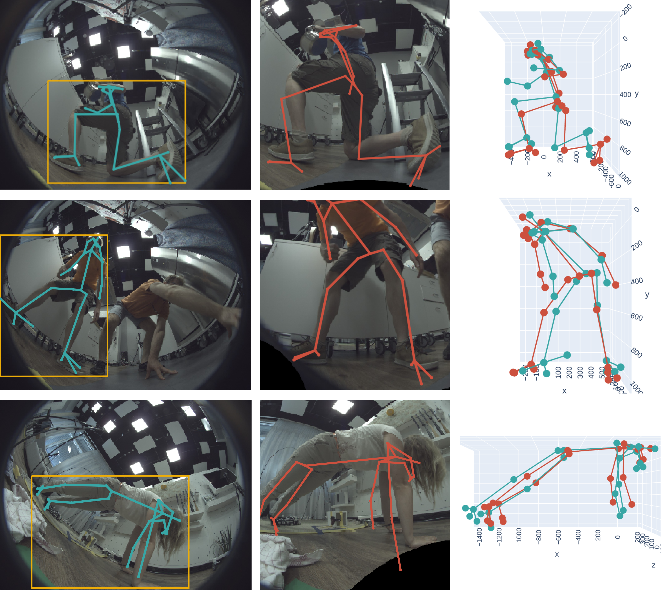} 
    \caption{Sample images from our \crazypose{} dataset, crops and predicted poses under heavy fisheye distortion.\\ Color legend: \textcolor[rgb]{0.219, 0.65, 0.647}{\rule{1ex}{1ex}} Ground truth
    \textcolor[rgb]{0.8, 0.3137, 0.2431}{\rule{1ex}{1ex}} Predictions}  
    \label{fig:example}
\end{figure}

Human pose estimation (HPE) is crucial for automotive systems \cite{Others:OmniDirectionalSurvey:Ai}, surveillance \cite{Applications:Surveillance:Cormier}, human-robot interaction \cite{HRI:HPE:Cheng}, action recognition \cite{Action:HDGCN:Lee}, and sports analysis \cite{Data:SportsPose:Ingwersen} \cite{Application:IndianDance:Jayanthi}. Fisheye cameras, with their wide field of view (FOV), capture extensive body movement and reduce the need for multiple cameras, lowering costs in robotics and surveillance. However, fisheye lenses introduce distortions, especially towards the image boundaries, which are not present with classic pinhole (PH) cameras.

Different solutions for handling fisheye images in HPE have been explored but lack systematic comparison \cite{Fisheye:Survey:Zhang}. Some of these methods reproject fisheye images to less distorted images \cite{Data:Zhang:3dhuman}, so that HPE models trained on regular PH images can be applied \cite{HPE:ISS:Minoda}. This paper presents the first comprehensive evaluation of reprojection models for monocular 3D HPE with fisheye images. Our comparison includes  reprojecting fisheye crops to PH format, using cylindrical barrel reprojections, applying fisheye camera models without reprojection as well as employing bounding box heuristics for optimal choice of projection. 
\begin{table}[h!]
\rowcolors{2}{white}{mygray}
    \centering
    \captionof{table}{Statistics of our novel \crazypose{} dataset.}
    \label{tab:dataset-stats}
    \resizebox{\linewidth}{!}{
    \begin{tabular}{lcccc}
        \toprule
        & \textbf{Living Room 1} & \textbf{Living Room 2} & \textbf{Kitchen} & \textbf{Total} \\
        \midrule
        \text{Subjects} & 4 & 5 & 7 & 7 \\
        \text{Subjects/Scene} & 1\text{--}2 & 1\text{--}3 & 1\text{--}4 & 1\text{--}4 \\
        \text{Sequences} & 13 & 22 & 21 & 56 \\
        \text{Images} &  &  &  &    \\
        ~\text{Fisheye} & 50,621 & 84,863 & 66,756 & 202,240 \\
        ~\text{Pinhole} & 51,407 & 42,532 & 44,359 & 138,298 \\
        \text{Camera Perspectives} &  &  &  &  \\
        ~\text{Fisheye} & 4 & 6 & 6 & 16 \\
        ~\text{Pinhole} & 4 & 3 & 4 & 11 \\
        \midrule
        \text{Fisheye Perspectives} & & & & \\
        ~\text{Horizontal $\rightarrow$} & \checkmark & \checkmark & \checkmark & \checkmark \\
        ~\text{Upwards $\uparrow$} & \checkmark & \checkmark & & \checkmark \\
        ~\text{Downwards $\downarrow$} & & & \checkmark & \checkmark \\
        ~\text{Tilted Downwards $\searrow$} & \checkmark & \checkmark & \checkmark & \checkmark \\
        ~\text{Tilted Upwards $\nwarrow$} & \checkmark & & \checkmark & \checkmark \\
        \bottomrule
    \end{tabular}
   } 
\end{table}

We extend the \textbf{MeTRAbs} \cite{HPE:Metrabs:Sarandi} 3D HPE method to support fisheye images by integrating both equidistant (ES) and generic double sphere (DS) camera models, marking the first integration of these models into an HPE framework. Our results show these models outperform traditional PH reprojection (without requiring retraining on fisheye data) especially for subjects covering large image areas or located close to the camera.

Additionally, we introduce a \textbf{new dataset called \crazypose{}} (Fisheye Imagery in Challenging Human Poses), designed to test HPE methods under unconventional camera angles and body poses.
Unlike existing fisheye HPE datasets (\cref{tab:fisheye-datasets}),
which are focused on head-worn, single-subject VR/AR applications or surveillance use-cases, our dataset targets robot-like perspectives with complex multi-person scenarios and close-up interactions.

Our main contributions are: (1) a systematic comparison of five projection models, (2) an extension of a state-of-the-art pose estimator with fisheye camera models, (3) a heuristic to dynamically choose the best projection model, and (4) a novel HPE evaluation dataset with challenging fisheye camera angles and unusual poses.

\begin{figure*}[h!]
  \centering
  \includegraphics[width=\textwidth]{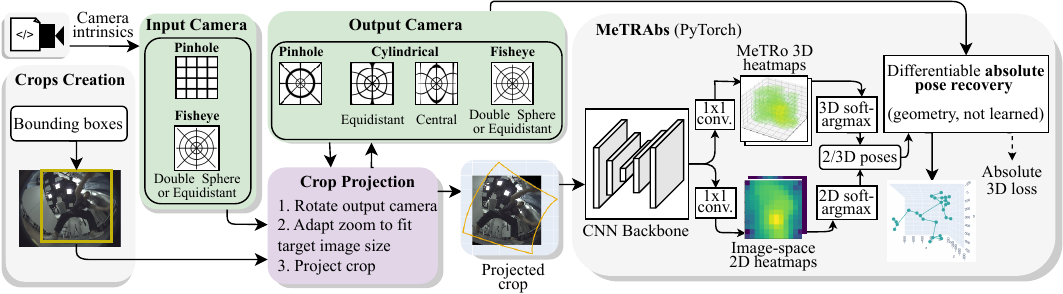}
  \caption{Our fisheye extension of \metrabs{}. The input camera is the camera that created the original image. Its intrinsics are required input for our method. The output camera can be chosen from 5 different models. We first create crops for each person in an image, then project the crop from the input camera to the output camera (see \cref{sec:out:pipeline}). The projected crop is then fed into \metrabs{}. Using the 2D prediction, relative 3D prediction and the camera parameters of the output camera, the absolute 3D pose is recovered by solving a system of linear equations.  More details in \cref{sec:out:pipeline}.}
  \label{fig:overview}
\end{figure*}

\section{Related Work}

\subsection{3D Human Pose Estimation on Pinhole Images} 
HPE is the process of locating the positions of key body joints of a person in an image or video. While some 3D HPE methods incorporate additional modalities like depth information \cite{HPE:KinectDepthBased:Shotton} or Inertial Measurement Unit data \cite{HPE:LidarAidInertialPoser:Ren}, we focus on top-down 3D HPE using monocular RGB images. Such methods \cite{HPE:TopDownSurvey:Nguyen, HPE:VirtualPose:Su}, including \metrabs{} \cite{HPE:Metrabs:Sarandi} and its extensions \cite{HPE:Autoencoder:Sarandi, HPE:MetrabsPlus:Matsune}, first detect persons in the image and then predict skeletons within detected bounding boxes.

3D HPE estimates joint coordinates either as absolute poses, relative to a global coordinate system, or as root-relative poses, defined with respect to a predefined root joint (e.g., pelvis) \cite{HPE:Metro:Sarandi, HPE:SMAP:Zhen, HPE:Metrabs:Sarandi}. For 3D keypoint estimation, approaches include direct 3D regression \cite{hpe:esraa:survey}, 2D heatmaps with 3D uplifting \cite{HPE:Lifting:Martinez}, and 3D heatmaps \cite{HPE:MonocularSurvey:Liu, HPE:TopDownSurvey:Nguyen}.

\subsection{Fisheye-based Human Pose Estimation}
Despite the prevalence of fisheye optics in autonomous driving \cite{Fisheye:SurroundViewSurvey_Driving:Kumar}, a recent survey by \cite{Fisheye:Survey:Zhang}  underlines that research on 3D HPE using fisheye images is limited. 
Related work mainly targets egocentric pose estimation for AR/VR \cite{HPE:Mo2Cap:Xu, HPE:EgoPose:Tome, Data:Rhodin:EgoCap, Data:Hakada:Unrealego, Data:Liu:EcHP}, surveillance with downward-facing cameras \cite{Data:Zhang:3dhuman, Data:ODIN:Ravin}, or person detection \cite{Fisheye:PeopleDetection_RaPiD:Duan, Data:WEPDTOF:Tezcan}. For human-robot interaction, horizontal below-eye-level perspectives \cite{HPE:ISS:Minoda, HPE:ChestMounted:Aso} and upward-facing views of robots are particularly relevant, as robots are often smaller than humans. However, to the best of our knowledge, upward-facing views have not been addressed in any study.

\subsubsection{Without Image Reprojection}
In AR/VR scenarios with head-mounted cameras \cite{HPE:Mo2Cap:Xu, HPE:EgoPose:Tome}, training HPE methods on raw fisheye images can be effective, as backbones can learn to manage distortion with a fixed camera setup. Some works use dedicated fisheye models like the omnidirectional model \cite{Camera:Omnidirectional:Scaramuzza}, as in \cite{Data:Rhodin:EgoCap}. \cite{Fisheye:Survey:Zhang} are the only ones to systematically evaluate several HPE and action recognition algorithms on a fisheye dataset.
They find that their own approach \cite{Data:Zhang:3dhuman}, the only baseline that explicitly incorporates a fisheye-specific (polynomial) camera model, yields best HPE results on their \textit{F-M3DHPE} dataset. It requires two separate backbones for relative and absolute pose recovery and is thus computationally more complex than our proposed method.

\subsubsection{Projection to Pinhole}
Reprojecting a fisheye image to PH format causes significant information loss \cite{Fisheye:Cylindrical:Plaut}. However, smaller sections can be projected if the FOV is below $120^\circ$–$140^\circ$. \cite{HPE:ISS:Minoda} applied this to a small dataset with fixed single horizontal camera setup but did not address cases with subjects close to the camera, where PH projection becomes infeasible.

\subsubsection{Projection to Cylinder Surface}
\textit{Plaut et al.} \cite{Fisheye:Cylindrical:Plaut, Fisheye:CylindricalTutorial:Plaut} found that cylindrical projections better preserve translation invariance for CNNs compared to spherical projections. \cite{HPE:ChestMounted:Aso} used cylindrical projections with chest-mounted, horizontal
fisheye cameras for 3D HPE.
\begin{table}[t]
\centering
\caption{Comparison of ego- and exocentric HPE/HAR datasets regarding fisheye camera orientation.}
\label{tab:fisheye-datasets}
\resizebox{\linewidth}{!}{%
\rowcolors{2}{white}{mygray}
\begin{tabular}{lcccccc}
    \toprule
    Dataset & Frames & Subj. & \makecell{Perspective \\ (fisheye only)}  & \faHome/\faSunO & Public & \makecell[l]{Real/ \\ Synth.} \\
    \midrule
    ODIN \cite{Data:ODIN:Ravin} & 332K & 15 &  exo: $\downarrow$ & \faHome & \checkmark & real \\
    CEPDOF \cite{Fisheye:PeopleDetection_RaPiD:Duan} &  25.5K & - &  exo: $\downarrow$ & \faHome & \checkmark & real \\
     3DhUman \cite{Data:Zhang:3dhuman} & 217 & 3 &  exo: $\downarrow$ & \faHome &  x & real \\
     OmniLab \cite{Data:Yu:Ntop} & 4.8K & 5 & exo: $\downarrow$ & \faHome & \checkmark & real \\
    $\rm Mo^2Cap^2$-train~\cite{HPE:Mo2Cap:Xu} & 530K & 700+ & ego & \faHome/\faSunO  & \checkmark & synth \\
    $\rm Mo^2Cap^2$-eval~\cite{HPE:Mo2Cap:Xu} & 5.6K & - & ego  & \faHome/\faSunO & \checkmark & real \\ 
    $x$R-EgoPose \cite{HPE:EgoPose:Tome} & 383K & 46 & ego  & \faHome/\faSunO & \checkmark & synth \\
    EgoCap \cite{Data:Rhodin:EgoCap} & 60K & 8 &  ego   & \faHome/\faSunO & \checkmark & real \\
    UnrealEgo \cite{Data:Hakada:Unrealego} & 900K & 17 & ego  & \faHome/\faSunO &  \checkmark & synth. \\
    ECHP \cite{Data:Liu:EcHP} & 92K & 11  & ego  & \faHome/\faSunO &  x & real\\
    EgoExo4D \cite{Data:EgoExo4D:Grauman} & 9.6M & 740 &  ego  & \faHome/\faSunO & \checkmark & real \\
    First2Third-Pose \cite{Data:Dhamanaskar:First2ThirdPose} & - & 14 & ego, exo: $\rightarrow \searrow$   & \faHome/\faSunO & \checkmark & real\\
    Nymeria \cite{Data:Nymeria:Ma} & 260M & 264 &  ego,  exo: $\rightarrow \searrow $    & \faHome/\faSunO & \checkmark & real \\      
    EgoHumans \cite{Data:EgoHumans:Khirodkar} & 125k & - &  ego, exo: $\rightarrow \searrow $  & \faHome/\faSunO &  \checkmark & real \\     
    F-M3DHPE \cite{Fisheye:Survey:Zhang} & 2.8K & 11 &  exo:  $\rightarrow \searrow$  & \faHome & \faEnvelopeO & real \\
    F-HAR \cite{Fisheye:Survey:Zhang} & - & 13 &  exo: $ \rightarrow \searrow$   & \faHome/\faSunO & \faEnvelopeO & real \\
    \midrule
    \rowcolor{white}
    \makecell[l]{\crazypose{}\emph{-F} (ours) \\ (fisheye subset, see \cref{tab:dataset-stats})} & 202K &  7 &  \makecell[l]{exo:
    $\downarrow \uparrow \nwarrow \searrow \rightarrow \& $ \\ extreme close-ups} & \faHome  & \checkmark & real\\
    \midrule
    \\
    \rowcolor{white}
    \multicolumn{7}{l}{
    Notation: indoor \faHome, outdoor \faSunO, on request \faEnvelopeO, exocentric fisheye camera orientation $\downarrow \uparrow \nwarrow \searrow \rightarrow$ }
\end{tabular}
}
\end{table}

\subsection{Public Datasets for Fisheye-Based HPE}
Most HPE and human action recognition (HAR) datasets use PH images \cite{Data:3DPW:Marcard, Data:H36M:Ionescu, Data:Mehta:MUPOTS}. Existing fisheye datasets (\cref{tab:fisheye-datasets}), mainly focus on egocentric or downward-facing surveillance views \cite{HPE:Mo2Cap:Xu, HPE:EgoPose:Tome, Data:Rhodin:EgoCap, Data:Hakada:Unrealego, Data:Liu:EcHP, Data:Zhang:3dhuman, Fisheye:PeopleDetection_RaPiD:Duan, Data:WEPDTOF:Tezcan, Data:ODIN:Ravin}.  Among them, \textit{Nymeria} \cite{Data:Nymeria:Ma} and \textit{EgoExo4D} \cite{Data:EgoExo4D:Grauman} use Meta's \textit{Project Aria Glasses}, which allow a fisheye-based human-perspective observation of the scene and other humans.
The datasets presented in \cite{HPE:ISS:Minoda} and \cite{HPE:ChestMounted:Aso} focus on third-person perspective and 
include horizontal views, but are not publicly accessible. 

Unlike previous works, we introduce the \crazypose{} dataset, comprising 202K images from diverse camera perspectives, including horizontal, floor-mounted cameras (tilted at $45^\circ$ or $90^\circ$), wall-mounted cameras angled downwards, various lens types, multi-person scenarios with occlusion, and a range of activities. It includes accurate 3D pseudo-ground truth obtained through multi-view triangulation from time-synchronized cameras. We assess various fisheye reprojection methods using the state-of-the-art \metrabs{} \cite{HPE:Metrabs:Sarandi} 3D HPE approach on our dataset.

\section{\metrabs{} Fisheye Extension}
The PH camera model is unsuitable for fisheye images due to its inability to account for spherical distortions that increase from center to edge. Fisheye lenses project a spherical view onto a flat image plane, causing straight lines to bend. Therefore, an HPE framework using fisheye input must incorporate camera models that represent these distortions.

We aim to assess the effectiveness of different fisheye reprojection methods on 3D HPE. To achieve this, we select the non-temporal, heatmap-based monocular HPE method \metrabs{} \cite{HPE:Metrabs:Sarandi} as baseline, reimplement it in PyTorch and extend it to enable handling of both PH and fisheye input images (see \cref{fig:overview}). We choose this method as it enables real-time performance even with constrained computational resources due to a lightweight backbone. Our implementation incorporates different camera models and reprojection methods to effectively manage fisheye distortions. Notably, the application of these (re-)\! projection methods only requires camera parameters of the fisheye optics without any further need for retraining. Our approach offers seamless flexibility for various fisheye models and is adaptable to any similar top-down HPE system.

\subsubsection{Fisheye Projection Methods}
\label{sec:Projections}
We implement two fisheye camera models: The \textbf{Equidistant Model (EF)} \cite{Camera:Equidistant:Hughes} assumes a direct proportionality between the angle from the optical axis and the radial distance from the image center. The \textbf{Double Sphere Model (DS)} \cite{Camera:DoubleSphere:Usenko} improves accuracy by projecting a 3D point onto two concentric spheres with shifted centers before mapping onto the image plane via a translated PH model. DS is better suited to model real fisheye lenses than the EF model. Unlike other fisheye models (e.\,g. polynomial) which may require an iterative approach for unprojection that is not easily usable in end-to-end training, the closed-form analytic inverse of DS facilitates obtaining normalized image coordinates for absolute pose recovery.

In contrast, a \textbf{cylindrical projection} flattens the curved surface onto a cylinder, mitigating radial distortion and preserving proportions more consistently over a $180^\circ$ field \cite{Fisheye:CylindricalTutorial:Plaut}. Our framework supports both \enquote{equidistant} (EC) and \enquote{central} cylindrical (CC) projections. 
In CC projection \cite{Fisheye:Cylindrical:Plaut}, rays from a central point intersect the cylindrical surface aligned with the Y-axis. This projection spans $360^\circ$ azimuthally but struggles near the cylinder axis, similar to PH camera limitations around $180^\circ$ \cite{Fisheye:CylindricalTutorial:Plaut}. The EC projection maps the vertical image coordinate to the polar angle instead of the cylinder axis, allowing it to represent the entire sphere.

\subsubsection{Original \metrabs} \textbf{MeTRAbs} \textbf{(Metric-Scale Truncation-Robust Heatmaps for Absolute 3D Human Pose Estimation)} \cite{HPE:Metrabs:Sarandi}, estimates an absolute 3D human pose given an image crop of a person and known input camera parameters. It feeds the person crop through a CNN backbone and outputs 3D as well as 2D image coordinates for the person's joints. Predicted 2D keypoints are then transformed to normalized image coordinates via the unprojection equation of the crop's camera model. From those, the absolute 3D pose is obtained by solving a strong perspective model via linear least squares~\cite{HPE:Metrabs:Sarandi}.

\subsubsection{Our Pipeline \& \metrabs{} Fisheye Extension} 
\label{sec:out:pipeline}
\cref{fig:overview} shows how we process PH/fisheye images.
Given an image of (multiple) persons, we apply an off-the-shelf person detector \cite{Detector:RTMDet:Lyu}. Backward warping is then applied to transform the resulting crop using one of the camera projections we intend to compare (see \ref{sec:Projections}). The projection applies the intrinsics of the input camera (the physical camera that created the image) and output camera (a virtual output camera with intrinsics chosen by the user). The latter is virtually pointed at the bounding box center, creating an output image which looks as if it was actually taken with the output camera.

The output cameras's zoom factor is chosen so that the centers of the sides of the bounding box lie within the resulting image. Afterwards, we feed the transformed crop into \metrabs{} and obtain the absolute 3D pose, using the output camera's intrinsics for the absolute pose recovery.

\subsubsection{Heuristics for Projection Choice}\label{Heuristic} 
\label{hybrid}
Our experiments (\ref{reprojection_methods_results}) show that for relative pose estimation, different projection types can be beneficial depending on the FOV covered by a person.  We thus introduce two metrics to estimate the FOV a person occupies (MPJA) and automatically select the most appropriate projection method for each image based on the spatial expansion of the bounding box (MBBA). 

\begin{figure}[h]
    \centering
    \begin{subfigure}[b]{0.45\linewidth}
        \centering
        \includegraphics[width=\linewidth]{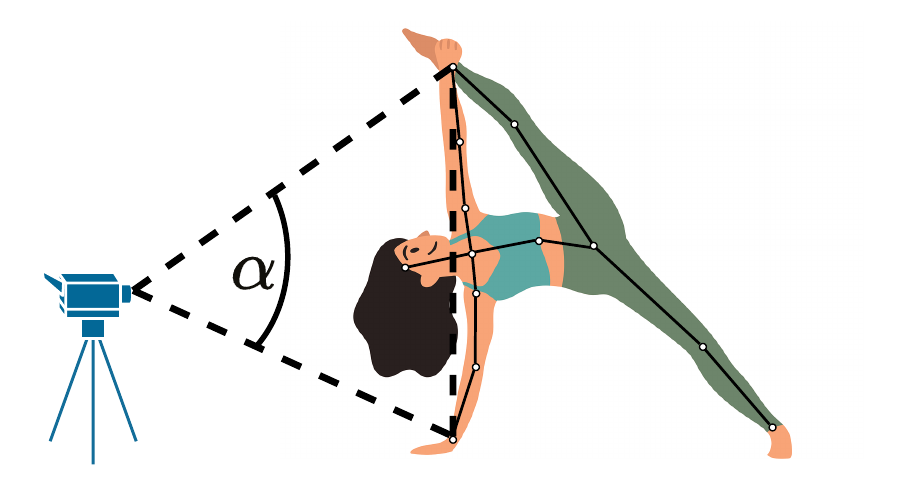}
        \caption{Example for MPJA ($\alpha$).}
        \label{fig:yoga-example-for-camera-angle}
    \end{subfigure}
    \hfill
    \begin{subfigure}[b]{0.45\linewidth}
        \centering
        \includegraphics[width=\linewidth]{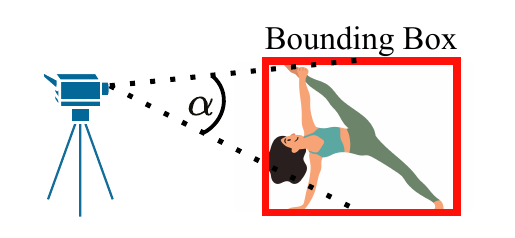}
        \caption{Example for MBBA ($\alpha$).}
        \label{fig:yoga-example-for-MPBBA}
    \end{subfigure}
    \begin{subfigure}[b]{0.45\linewidth}
        \centering
        \includegraphics[width=\linewidth, height=25mm]{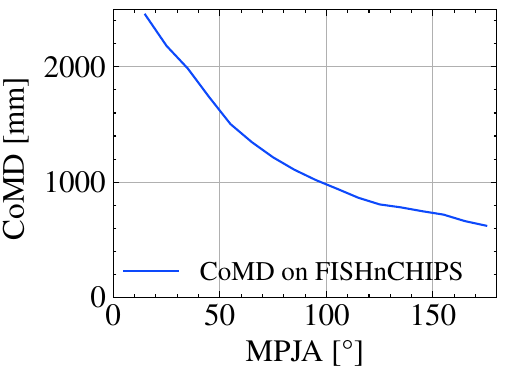}
        \caption{Dist. [\unit{mm}] vs. MPJA [$^\circ$].}
        \label{fig:distance}
    \end{subfigure}
    \hfill
    \begin{subfigure}[b]{0.42\linewidth}
        \centering
        \includegraphics[width=\linewidth, height=25mm]{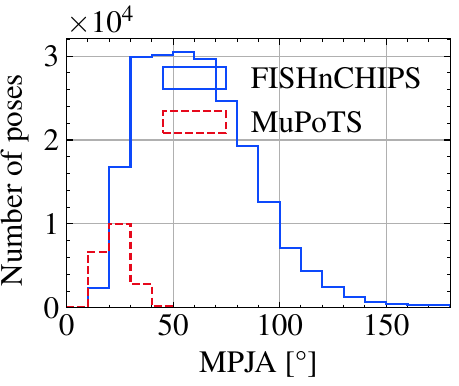}
        \caption{MPJA Histograms.}
        \label{fig:histogram}
    \end{subfigure}    
    \caption{Top row: Our novel spatial expansion metrics. MPJA (a) takes the maximum pairwise angle of joints. MBBA (b) computes the maximum opening angle of the (reprojected) bounding box. Bottom row: (c) Plot depicts correlation between a skeleton's CoM-camera-distance (CoMD) and MPJA. (d) Difference in MPJA counts for MuPoTS-3D-val \cite{Data:Mehta:MUPOTS} and our \crazypose{} dataset. \textit{(Yoga pose credit: \cite{Others:YogaImage:Pixabay}})}
    \label{fig:combined_figure}
\end{figure}

(a) Analysis Tool: The \textbf{\emph{Maximum Pairwise Joint Angle (MPJA)}} calculates the maximum angular difference for all joint pairs in a human's pose using the ground truth skeleton (\cref{fig:yoga-example-for-camera-angle}). MPJA helps quantifying fisheye distortions of a person by distinguishing between poses that occupy a small FOV (suitable for PH projection) from those spanning a larger FOV. It allows to reveal limitations of the PH projection model as seen in our experiments (Figure~\ref{fig:angular_pose_results_plots}).

(b) Prediction Tool:
Calculating MPJA requires ground truth poses, which are unavailable during inference. To maximize HPE accuracy, we rely solely on the person’s bounding box coordinates to estimate the FOV. After projecting the 2D bounding box into 3D, we calculate the angles relative to the camera center and select the maximum angle (\textbf{\emph{Maximum Bounding Box Angle, MBBA}}), as shown in \cref{fig:yoga-example-for-MPBBA}. The optimal projection technique is selected based on MBBA during inference. We define a threshold $\alpha_t$ for MBBA and suggest to apply PH projection below the threshold and DS above it, resulting in a  \textbf{hybrid projection method} (\textbf{H}). Our experiments (\cref{tab:eval:crazypose:fisheye}) demonstrate that MBBA closely approximates MPJA, yielding nearly identical inference results.

\section{Novel Fisheye Evaluation Dataset}
\label{sec:our-dataset}
For evaluation, we introduce \crazypose{}, a novel multi-view fisheye dataset, tailored to household robotics scenarios (living room, kitchen), unlike existing datasets focused on AR/VR and surveillance applications. The dataset counts more than 200,000 fisheye images and 138,000 PH (Kinect) images, with fisheye images captured from four different camera orientations (upwards, downwards, angled, and horizontal). It also includes near-floor camera placements tilted at $45^\circ$ and $90^\circ$ upwards, which are particularly relevant for low-height domestic service robots. The dataset features diverse indoor environments, clothing styles, textures, and lighting conditions. Seven subjects perform activities such as embracing, yoga, unloading boxes, climbing ladders and robot navigation gestures. The distribution of MPJA values is shown in \cref{fig:histogram} as histogram and compared to the MuPoTS dataset \cite{Data:Mehta:MUPOTS}; 5500 images capture poses with MPJA$>\!120^\circ$.

\subsubsection{Camera Setup} Data was recorded using 10 synchronized cameras at 15 Hz and later downsampled to 5 Hz: 6 machine vision color cameras with fisheye lenses (\qty{2.4}{MP} and \qty{5.1}{MP}) and 4 Azure Kinect RGB-D cameras (\qty{3.1}{MP}) with PH characteristics. We used S-mount and C-mount fisheye lenses with varying costs (\euro{4} to \euro{800}) and distortion levels. Calibration was performed using the DS model for fisheye cameras and OpenCV’s PH model for Kinects, with extrinsic calibration via commercial software (cf. \cite{Others:MultiviewCalibration:Linder}).

\subsubsection{Ground Truth} We obtain precise 3D GT through multi-view triangulation across all 10 camera views. Accurate 2D bounding box detection is achieved using Co-DINO-DETR \cite{Detector:CO-DETR:Zong} with Swin-L \cite{Detector:SWIN:Liu} backbone, followed by RTMPose-L trained on 7 public datasets~\cite{HPE:RTMPose:Jiang} for 2D pose estimation. In multi-person scenarios, we associate 2D skeletons of the same human across all views based on distances to projected, pre-recorded Kinect 3D skeletons. Triangulation is conducted using respective camera models with a non-linear least-squares solver \cite{Others:CeresSolver:Agarwal}, a bone symmetry constraint, and several human anatomy-based skeleton plausibility checks.

Statistics of our dataset are provided in \cref{tab:dataset-stats}, with examples in \cref{fig:example}. \crazypose{} features 7 subjects in 3 setups, with a total of 56 video sequences. Overall, it contains approximately 340,000 images from 16 fisheye and 11 PH camera perspectives. For more details, please refer to our supplementary video. An anonymized version of the dataset with blurred faces will be released on our project website.
\section{Implementation \& Training}
For our HPE framework, we re-implemented the newest \metrabs{} version \cite{HPE:Autoencoder:Sarandi} which uses an autoencoder to allow training on datasets with different skeleton formats. We use the consistency-finetuning training variant presented in  \cite{HPE:Autoencoder:Sarandi} on an EfficientNetV2-S \cite{Others:EfficientNetV2:Tan} backbone, using the same model for base training and autoencoder-based finetuning. 
We use \textit{Sárándi et al.}'s collection of 28 PH datasets \cite{HPE:Autoencoder:Sarandi} as training data and compare our predictions to the original \metrabs{} paper on \textit{MuPoTS-3D} \cite{Data:Mehta:MUPOTS} with respect to the Percentage of Correct Keypoints (PCK), as this metric was the most common among recent methods. As shown in \cref{tab:baseline_mupots}, our re-implementation of \metrabs{} is still within the current state of the art considering that we are using a lightweight backbone and small crop resolution of 256$\times$256 \unit{px}.

We do not require any retraining of the core \metrabs{} to use fisheye input images. However, the joint annotation schema for our novel dataset is different from any in the training data. We thus need to apply a mapping between training skeleton formats and our skeleton format. To this end, we adopt the autoencoder proposed in \cite{HPE:Autoencoder:Sarandi} and train a linear transformation on a separate held-out sub-dataset of \crazypose{} (21k images) recorded with different human subjects in an another room, but with a similar camera setup as described in \cref{sec:our-dataset}.

\begin{table}[t]
\rowcolors{2}{white}{mygray}
    \centering
    \vspace*{2mm}
    \caption{Comparison of our MeTRAbs baseline to other recent methods on MuPoTs-3D \cite{Data:Mehta:MUPOTS} (pinhole images, no temporal information used).
    }
    \label{tab:baseline_mupots}
    \resizebox{\linewidth}{!}{
        \begin{tabular}{p{23mm}lp{21mm}cc} 
            \toprule
             \multirow{2}{*}{Method} &  \multirow{2}{*}{Backbone} & \multirow{2}{*}{Training Data} & \multicolumn{2}{c}{PCK\textsubscript{150} [\unit{\percent}]} \\
            \cmidrule(lr){4-5}
            & & & Abs.$\uparrow$ & Rel. $\uparrow$ \\
        \midrule
        \rowcolors{2}{mygray!20!white}{white}
            \textbf{Dual Network} \cite{HPE:DualNetwork:Cheng} & HRNet-w32 \cite{HPE:HRNet:Sun} & MuCo\,\cite{Data:Mehta:MUPOTS}\,+ COCO \cite{Data:Lin:COCO} & 48.1 & 89.6 \\
            \textbf{PIRN} \cite{HPE:permutationinvariant:Ugrinovic} & RootNet \cite{HPE:RootNet:Moon} & MuPoTs\,\cite{Data:Mehta:MUPOTS} cross-validation &  44.1 & 85.8     \\
            \textbf{VirtualPose} \cite{HPE:VirtualPose:Su} & ResNet-18/-152 \cite{Others:ResNet:He} & MuCo \cite{Data:Mehta:MUPOTS} & 44.0 & -    \\ 
            \textbf{GR-M3D} \cite{HPE:GR-M3D:Qiu} & Hourglass \cite{HPE:StackedHourglass:Newel} & MuCo \cite{Data:Mehta:MUPOTS} & 41.2 & 84.6\\
        \midrule
            \textbf{MeTRAbs} & ResNet-50 \cite{Others:ResNet:He} & MuCo \cite{Data:Mehta:MUPOTS} & 40.2  & 81.1 \\
             \textbf{MeTRAbs} + AE & EffNetV2-L \cite{Others:EfficientNetV2:Tan} & 28 Datasets & - & \textbf{95.4} \\
            \textbf{MeTRAbs} + AE & EffNetV2-S \cite{Others:EfficientNetV2:Tan} & 28 Datasets & - & 94.9 \\
        \midrule
            \textbf{MeTRAbs} (our re-implementation) & EffNetV2-S \cite{Others:EfficientNetV2:Tan} & 28 Datasets & \textbf{59.5} & 89.4 \\
            \bottomrule
            AE = Autoencoder, & as in \cite{HPE:Autoencoder:Sarandi} & & &  \\
        \end{tabular}
    }
\end{table}
\section{Metrics \& Experiments} 
\label{compare_projections}
\begin{figure*}[tb!]
    \captionsetup[subfigure]{justification=centering,margin={0.55cm,0cm}}
    \centering
    \begin{subfigure}{0.42\linewidth}
        \centering
        \includegraphics[width=\linewidth]{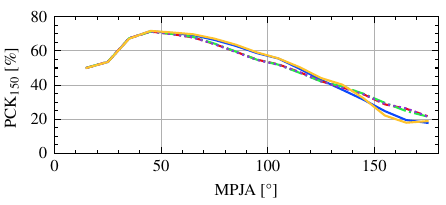}
        \caption{PCK$_{150}$$\uparrow$}
        \label{fig:pck}
    \end{subfigure}%
    \hfill
    \begin{subfigure}{0.42\linewidth}
        \centering
        \includegraphics[width=\linewidth]{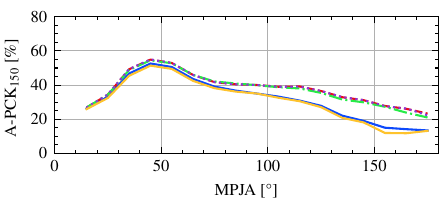}
        \caption{A-PCK$_{150}$$\uparrow$}
        \label{fig:apck}
    \end{subfigure}
    
    \vspace{0.5cm}

    \begin{subfigure}{0.32\linewidth}
        \centering
        \includegraphics[width=\linewidth]{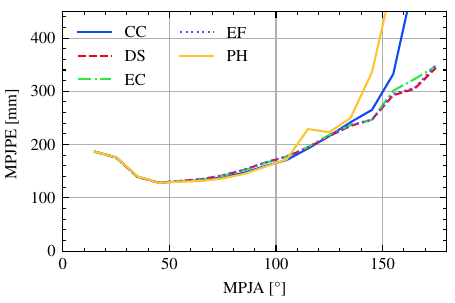}
        \caption{MPJPE$\downarrow$}
        \label{fig:mpjpe}
    \end{subfigure}%
    \hfill
    \begin{subfigure}{0.32\linewidth}
        \centering
        \includegraphics[width=\linewidth]{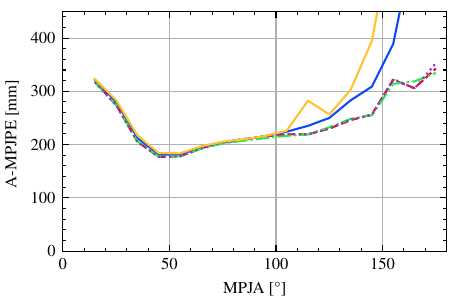}
        \caption{A-MPJPE$\downarrow$}
        \label{fig:ampjpe}
    \end{subfigure}
    \hfill
    \begin{subfigure}{0.32\linewidth}
        \centering
        \includegraphics[width=\linewidth]{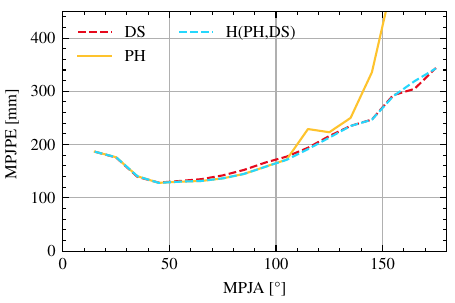}
        \caption{MPJPE$\downarrow$ (H-model)}
        \label{fig:select_hybrid}
    \end{subfigure}   
    \caption{MPJPE/PCK and A-MPJPE/A-PCK results depending on MPJA, evaluated on our fisheye dataset \crazypose{} for different projection methods. Measured by MPJPE and PCK, Pinhole projection (PH) performs best for poses within pinhole-like FOVs ($<120^\circ$), whereas fisheye camera projections (DS, EF) and EC-projection are more suitable for large FOVs ($>120^\circ$). DS and EF perform extremely similarly. \cref{fig:select_hybrid} depicts that the H-model combines the benefits from PH and DS: it performs identical to PH between $50^\circ$ and $110^\circ$ and comparable to DS for $110^\circ$--$150^\circ$.}
    \label{fig:angular_pose_results_plots}
\end{figure*}
Building on our \metrabs{} fisheye extension, we explore best practices for HPE on fisheye images by evaluating our novel spatial expansion metrics, systematically comparing (re-)projection methods, and testing our hybrid reprojection approach.

To investigate the reprojection method's effects on absolute and relative human pose reconstruction quality, we evaluate our results with respect to the HPE metrics \textbf{Percentage of Correct Keypoints (PCK)} and \textbf{Mean Per Joint Position Error (MPJPE)} as well as their counterparts for absolute poses (A-PCK, A-MPJPE), see \cite{HPE:Thesis:Sarandi} for details. To ensure detector-independent results, all experiments are conducted using GT bounding boxes. In real-world usage, they would be provided by a person detector.

\subsection{Experiments on Spatial Expansion Metrics}

Fisheye images distort individuals close to the camera or with poses covering a wide FOV. To validate MPJA as a heuristic for wide FOV coverage, assumed to correlate with higher distortion, we visually assess the relationship between a person’s FOV coverage and proximity to the camera. This is done by plotting MPJA against the \textbf{Center of Mass Distance to Camera (CoMD)}, which we define as the mean Euclidean distance from each joint to the camera center.

\subsection{Experiments on Reprojection Methods}
We used our \metrabs{} extension to systematically evaluate various reprojection methods on the \crazypose{} dataset by reprojecting cropped fisheye images to:  

\begin{itemize}
    \item \textbf{PH}: Pinhole camera model.
    \item \textbf{EF}: Equidistant fisheye model.
    \item \textbf{DS}: Double sphere model.
    \item \textbf{CC \& EC}: Cylindrical projections, distinguishing between central and equidistant cylindrical models.
    \item \textbf{H}: Our MBBA-threshold-based hybrid method, dynamically selecting PH or DS, based on the estimated FOV covered by the person.
\end{itemize}

We evaluate these methods by plotting PCK and MPJPE against MPJA to analyze their suitability for poses spanning different FOVs.
\section{Results \& Interpretation}
\subsection{Experimental Results on Spatial Expansion Metrics}
Our assumption was that subjects close to the camera (less than 1\,m away) would cover a large FOV in the image. Likewise, small FOV coverage would correspond to very distant or crouched poses. This assumption is supported by our data: a plot of CoMD against the MPJA shows that low MPJA values correspond to high CoMD and vice versa (see \cref{fig:distance}). This insight helps interpret the results of our comparison of different fisheye reprojection methods. \textit{(Note: All MPJA plots bin the data in $10^\circ$-intervals.)}

\subsection{Experimental Results on Reprojection Methods}
\label{reprojection_methods_results}
\begin{table}[!b]
    \centering
    \vspace*{3mm}
    \caption{(A-)MPJPE [mm] and (A-)PCK$_{150}$ [\%] results on our \crazypose{} fisheye 
    dataset. H stands for the proposed H projection, which uses a heuristic based upon MPJA/MBBA thresholds $\alpha_{t}$ to dynamically switch between PH and DS projection models. \textbf{Bold}: best, \underline{underlined}: 2nd.}
    \label{tab:eval:crazypose:fisheye}
    \resizebox{\linewidth}{!}{
    \begin{tabular}{lcccc}
         \toprule         
         Projection &  MPJPE$\downarrow$ & A-MPJPE$\downarrow$ & PCK$_{150}\uparrow$ & A-PCK$_{150}\uparrow$  \\
         \midrule
          CC  & 146.5 & 209.9 & 64.8 & 42.5 \\
           \cellcolor{mygray}DS  & \cellcolor{mygray}147.7 & \cellcolor{mygray}\underline{205.2} & \cellcolor{mygray}63.6 & \cellcolor{mygray}\textbf{45.5} \\
           EC  & 147.4 & \textbf{204.4} & 63.7 & \underline{45.4}\\
           \cellcolor{mygray}PH  & \cellcolor{mygray}151.6 & \cellcolor{mygray}219.4 & \cellcolor{mygray}\underline{65.1} & \cellcolor{mygray}41.5 \\
           EF & 147.8 & 205.3 & 63.5 & \textbf{45.5} \\
           \cellcolor{mygray}H w/MPJA & \cellcolor{mygray} & \cellcolor{mygray} & \cellcolor{mygray} & \cellcolor{mygray} \\
           \cellcolor{mygray}~$\alpha_{t}=110^\circ$& \cellcolor{mygray}\textbf{145.1} & \cellcolor{mygray}210.4 & \cellcolor{mygray}\underline{65.1} & \cellcolor{mygray}42.0\\
           \cellcolor{mygray}~$\alpha_{t}=135^\circ$& \cellcolor{mygray} 145.9&  
           \cellcolor{mygray}212.1& \cellcolor{mygray}\textbf{65.2} & \cellcolor{mygray}41.7\\
           H w/MBBA & & & & \\
           ~$\alpha_{t}=110^\circ$ & \textbf{145.1} & 210.2 & \underline{65.1} & 42.0\\
           ~$\alpha_{t}=135^\circ$ & \underline{145.8} & 211.8 & \textbf{65.2} & 41.7\\
           
          \bottomrule
    \end{tabular}%
    }
\end{table}

To compare the quality of the projection methods, we assess both the overall (A-)MPJPE and (A-)PCK on our full \crazypose{} dataset.
As seen from \cref{tab:eval:crazypose:fisheye}, for absolute pose estimation, using DS as output camera slightly outperforms other methods as it has the highest A-PCK (\qty{45.5}{\percent}) and second-lowest A-MPJPE (\qty{205.2}{mm}). For relative pose estimation, no single best practice emerges. 

To evaluate each projection's applicability to handle various amounts of FOV covered by the depicted person, we analyse MPJPE and PCK metrics against MPJA. \cref{fig:angular_pose_results_plots} reveals that for relative pose estimation, reprojection method performance varies with the MPJA of the pose:

(1) \textbf{MPJA $\mathbf{<50^\circ}$}: No projection method shows clear dominance. Small MPJA values indicate a small FOV covered by the pose. Thus, in this MPJA interval, we assume minimal fisheye distortion and therefore less pronounced differences among projections. 

(2) \textbf{$\mathbf{50^\circ<}$ MPJA $\mathbf{<150^\circ}$}: In this MPJA range, PH projection slightly outperforms others in relative pose estimation (\cref{fig:mpjpe}), as the FOV covered by the person remains within PH's effective range, minimizing pinhole reprojection artifacts. As shown in \cref{fig:mpjpe}, poses with MPJA between $40^\circ$ and $70^\circ$ yield optimal results, likely because their reprojections closely resemble the model's pinhole training data. MPJPE increases from \qty{115}{mm} at $50^\circ$ MPJA to over \qty{250}{mm} at $150^\circ$, reflecting the growing spherical distortion. 
For absolute pose estimation, however, PH projection does not outperform other methods in this range, as indicated by A-MPJPE and A-PCK metrics.

(3) \textbf{MPJA $\mathbf{>120^\circ}$}: DS and EF projections achieve the best absolute and relative pose predictions, followed by EC. At high MPJA, PH projection's MPJPE rises rapidly, likely because the FOV exceeds PH’s effective range, causing detail loss due to the small zoom factor during reprojection. In contrast, errors in other projections increase more gradually, as they are designed for larger FOVs.

As seen in \cref{fig:angular_pose_results_plots}a-d, CC projection underperforms compared to EC, DS, and EF, likely due to its limited $180^\circ$ vertical FOV (\cref{sec:Projections}), restricting its effectiveness across different camera orientations. Generally, cylindrical projections, constrained by their symmetry, are less versatile than spherical models. DS and EF appear to better handle diverse camera orientations, which may explain their slight advantage over EC.

We conclude that, for large FOV poses (MPJA $> 120$°), DS, EC, and EF projections are preferred over PH and CC. For smaller FOVs, PH projection is recommended for relative pose estimation. For absolute pose estimation, using DS is optimal irrespective of the pose's FOV coverage. 

\subsection{Results on Predicting the Best Projection Model} 
\cref{reprojection_methods_results} illustrates the benefits of selecting projections based on MPJA. As described in \cref{Heuristic}, we developed a hybrid method (H) that uses an MBBA threshold $\alpha_t$ to switch projections: PH for MBBA  $<\alpha_t$ and DS for MBBA $>\alpha_t$. In \cref{tab:eval:crazypose:fisheye}, we exemplary depict results for $\alpha_t \in \{110^\circ, 135^\circ\}$, as these yielded the best MPJPE and PCK, respectively. As expected, these values align with the PH-DS curve intersections in \cref{fig:pck} and \cref{fig:mpjpe}.

Additionally, we compute the heuristic using MPJA from GT joints. The H-method results  in \cref{fig:select_hybrid} confirm that the H-curve follows PH below $\alpha_t=110^\circ$ and DS above it. The heuristic improves MPJPE by \qty{2}{mm} over using a single PH/DS projection, demonstrating its effectiveness for relative pose estimation by leveraging both projections’ strengths. As expected, A-MPJPE drops by \qty{6}{mm}, since PH remains the weakest model for absolute pose estimation (\cref{fig:ampjpe}, \cref{fig:apck}), independent of MPJA or MBBA.
\section{Conclusions}
In this paper, we have explored the integration of various
fisheye camera and reprojection models into a state-of-the-art
3D HPE framework.
Our novel \crazypose{} dataset, with focus on challenging camera perspectives encountered in robotics use-cases, allowed us to assess these projection models on close-up interaction scenarios where the person covers a significant part of the camera's FOV. Here, we found that PH and CC projections deliver suboptimal performance.
In contrast, EF and DS projections performed best, with EC projections performing slightly worse.
We also introduced a heuristic for dynamically selecting a suitable  
projection model based on estimated bounding box geometry.

In summary, we believe that our experiments shed valuable insights into how fisheye cameras can successfully be used for 3D HPE in various human-robot interaction scenarios.

\FloatBarrier
\section*{Acknowledgment}
This work was funded by Robert Bosch GmbH under the project ``Context Understanding for Autonomous Systems''. The development of the multi-view data recording setup was supported by the EU Horizon 2020 research and innovation program under grant agreement 101017274 (DARKO).

\bibliographystyle{IEEEtran}
\bibliography{
    IEEEabrv,
    ConverenceAbrv,
    references/references_camera_models,
    references/references_CV_basics,
    references/references_data,
    references/references_detectors,
    references/references_fisheye,
    references/references_hpe,
    references/references_others
}

\addtolength{\textheight}{-12cm}

\end{document}